\def\year{2017}
\documentclass[letterpaper]{article}
\usepackage{aaai17}
\usepackage{times}
\usepackage{helvet}
\usepackage{courier}
\usepackage{graphicx}
\usepackage{amssymb}
\usepackage{amsmath}
\usepackage[linesnumbered]{algorithm2e}
\usepackage{mdwlist}
\usepackage{url}
\usepackage{ wasysym }
\usepackage{setspace}
\usepackage{color}
\usepackage{textcomp}
\usepackage{booktabs}
\usepackage{ccicons, xcolor, soul}
\usepackage{wrapfig}
\usepackage{subcaption}
\usepackage{sidecap}
\usepackage{multirow}
\usepackage{hhline}

\makeatletter
\newcommand{\xRightarrow}[2][]{\ext@arrow 0359\Rightarrowfill@{#1}{#2}}
\makeatother

\begin{document}
\title{Tweeting AI: Perceptions of Lay vs Expert Twitterati}
\author{Lydia Manikonda, Subbarao Kambhampati \\ Arizona State University\\ \tt \{lmanikonda, rao\}@asu.edu } 

\nocopyright

\maketitle
\begin{abstract}
\begin{quote}
With the recent advancements in Artificial Intelligence (AI), various organizations and individuals are debating about the progress of AI as a blessing or a curse for the future of the society. This paper conducts an investigation on how the public perceives the progress of AI by utilizing the data shared on Twitter. Specifically, this paper performs a comparative analysis on the understanding of users belonging to two categories -- general AI-Tweeters (AIT) and expert AI-Tweeters (EAIT) who share posts about AI on Twitter. Our analysis revealed that users from both the categories express distinct emotions and interests towards AI. Users from both the categories regard AI as positive and are optimistic about the progress of AI but the experts are more negative than the general AI-Tweeters. Expert AI-Tweeters share relatively large percentage of tweets about their personal news compared to technical aspects of AI. However, the effects of automation on the future are of primary concern to AIT than to EAIT. When the expert category is sub-categorized, the emotion analysis revealed that students and industry professionals have more insights in their tweets about AI than academicians. 

%Characterization of users manifested that `London' is the popular location of users from where they tweet about AI. Tweets posted by AIT are highly retweeted than posts made by EAIT that reveals greater diffusion of information from AIT.
\end{quote}
\end{abstract}

\section{Introduction}
Due to the rapid progress in the field of AI, there have been widespread various discussions and concerns about the threats and benefits of AI. These discussions include -- improving the everyday lives of individuals (https://goo.gl/ViLdgV), ethical issues associated with the intelligent systems (https://goo.gl/5KlmXk), etc. Social media platforms are ideal repositories of opinions and discussion threads. Twitter is one such popular platform where individuals post their statuses, opinions and perceptions about the ongoing issues in the society~\cite{Java2007,Naaman2010,yang2010predicting}. Given the ease of finding individual's perceptions and opinions on this platform, this paper  investigates the posts about AI shared on Twitter. Specifically, we compare and contrast the perceptions of users from two categories -- general AI-Tweeters (AIT) and expert AI-Tweeters (EAIT). We believe that the findings from this analysis can help research funding agencies, organizations, industries, and especially the AAAI community about the public perceptions of AI. 

There have been earlier efforts to understand public perceptions of AI. Recent work by Fast et. al~\cite{EthanAAAI2016} conducts a longitudinal study of articles published on AI in New York Times between January 1986 and May 2016 have evolved over the years. This study revealed that from 2009 the discussion on AI has sharply increased and is more optimistic than pessimistic. It also found that fears about losing control over AI systems have been increasing in the recent years. Another recent survey~\cite{LeslieHarvard2016} conducted by the Harvard Business Review on individuals who do not have any background in technology, stated the positive perceptions of these individuals toward AI. In contrast, despite online social media platforms being the main channels for communicating personal opinions~\cite{Java2007,Naaman2010,yang2010predicting}, there is no existing work on how users think and what users share about AI on these platforms. We aim to analyze the perceptions of individuals as manifested in their posts shared on Twitter. 

Towards this goal, we attempt to answer 5 important questions through a thorough quantitative and comparative investigation of the posts shared on Twitter. 1) What are the insights that could be learned by characterizing the individuals and their interests who are making AI-related tweets? 2) What is the Twitter engagement rate for the AI-tweets? 3) Are the posts about AI optimistic or pessimistic? 4) What are the most interesting topics of discussion about AI to the users? 5) What can we learn about the trends of semantic co-occurrences about AI? We address each of these questions in the next few sections. 

Our analysis reveals intriguing differences between the posts shared by AIT and EAIT on Twitter. Specifically, this analysis reveals five interesting findings about the perceptions of individuals who are using Twitter to share their opinions about AI. Firstly, users from both the categories are emotionally positive (or optimistic) towards the progress of AI. Secondly, even though users are positive overall, {\em expert users are more negative than AIT}. Thirdly, from the topics extracted from the tweets, expert users share large percentage of tweets about their personal news. Fourth, the effects of automation on the future are of predominant concern to AIT. Lastly, when we sub-categorized EAIT, the emotional analysis on AI-related tweets revealed that {\em academicians are less positive and more social than students and industry professionals}. 

%we found that the tweets shared by EAIT have lower diffusion rate than the tweets posted by AIT as measured by the magnitude of retweets. Fourth, and last, users from AIT are geographically distributed (mostly Europe and US) with London and New York City taking the top positions.

%Some of the other findings from this analysis includes the popular topics of interest to users from AIT and EAIT. When examined the co-occuring patterns of AI vocabulary, it revealed that EAIT thinks deeper than AIT. 
In the next section, we describe the process of collecting data from the two categories of users considered in this paper. There after, we compare and contrast the perceptions of AIT and EAIT on each of the 5 research questions we posed earlier. We then present the summarization of our findings from the investigation presented in this paper in conclusions.

%Our analysis revealed intriguing differences between the way Twitter users and experts share their opinions and engage themselves in discussions about AI. The posts made by experts are more likely to be favorited whereas the posts made by Twitter users are more likely to be retweeted. Existing literature~\cite{Kwak2010,yang2010predicting} considers retweets as an important metric to measure the gravity of information diffusion. This result might suggest that the information spread among the Twitter users is relatively high compared to the experts. The $n$-gram analysis and topic analysis suggest that experts predominantly talk about AI. Also, experts consider sharing news and research-related information relatively higher than sharing their opinions which we notice among the Twitter users. 

%The analysis that extracts co-occurring concepts to measure the perceptions of individuals revealed that Twitter users engage themselves in discussions about AI as if the advancements envisioned are already happening. However, experts focus on discussing about the possibilities of the envisioned theories happening in the future. However, both these sets of users believe that the benefits of AI are good and their emotional gravities are leaning more towards being positive or optimistic. It is surprising to notice that the tweets related to AI are positive in contrast to general tweets~\cite{LeslieHarvard2016,LydiaICWSM2016}. We attach an appendix to this paper that presents the similar but extended analysis on Reddit to examine and compare with the findings obtained from Twitter. 

\section{Data Collection}
%To assess the perceptions about AI manifested in social media, we utilize the data from the popular microblogging platform -- Twitter (https://twitter.com/). A survey by Pew Research\footnote{http://www.pewinternet.org/fact-sheet/social-media/} states that Twitter is one of the prominent sources of communication. Recent study by Manikonda et al.~\cite{LydiaICWSM2016} shows that Twitter is primarily used to share higher percentages of posts that are about work or reviews of products available in the market. This is one of the motivations to consider tweets to analyze the public perceptions of AI.
% as a primary platform to crawl the tweets focusing on artificial intelligence and its related topics. 

\subsection{AI-Tweeters (AIT)}
We employ the official API of Twitter\footnote{https://dev.twitter.com/overview/api} along with a frequency-based hashtag selection approach to crawl data from the general Twitter users who tweet about AI. We first identify an appropriate set of hashtags that focus on the artificial intelligence in social media. In our case, these are -- \#ai and \#artificialintelligence. With these seed hashtags, we crawled 2 million unique tweets and then iteratively extracted hashtags to identify the most frequent co-occurring hashtags with the seed set. We then remove non-technical hashtags (for example: \#trump, \#politics, etc) from this sorted hashtag list. The top-15 co-occurring hashtags after this pre-processing are shown in Table~\ref{tab:cooccurringhts}. 
%To collect the dataset for Twitter users, we utilize a frequency-based hashtag selection approach to crawl the dataset. To perform this crawling, we employ Twitter official API~\footnote{https://dev.twitter.com/overview/api}. We first identify an appropriate set of hashtags that focus on the artificial intelligence in social media which in our case are -- \#ai and \#artificialintelligence. With these seed hashtags, we crawl 2 million unique tweets. By considering these tweets, we iteratively compute the co-occurring hashtags and sort them based on their frequency. We remove non-technical hashtags included in this sorted hashtag list for example: \#trump, \#politics, etc. The top-15 co-occurring hashtags after the pre-processing are shown in Table~\ref{tab:cooccurringhts}. 

\begin{table}[h]
\small
\centering
\setlength{\tabcolsep}{3pt}
\begin{tabular}{l l l} \hline
1. \#ai & 2. \#artificialintelligence & 3. \#machinelearning \\
4. \#bigdata & 5. \#iot & 6. \#deeplearning \\
7. \#robotics & 8. \#datascience & 9. \#cybersecurity \\
10. \#vr & 11. \#ar & 12. \#nlp \\
13. \#ux & 14. \#algorithms  & 15. \#socialmedia \\ \hline
\end{tabular}
\caption{Top-15 co-occurring hashtags with the seed hashtags: \#ai and \#artificialintelligence}
\label{tab:cooccurringhts}
\end{table}

We used the top-4 hashtags from this list: \emph{\#ai}, \emph{\#artificialintelligence}, \emph{\#machinelearning} and \emph{\#bigdata} as the final hashtag set to crawl a set of 2.3 million tweets. We found that the set of tweets obtained using these 4 hashtags are a superset of all the tweets crawled by utilizing the remaining hashtags presented in Table~\ref{tab:cooccurringhts}. Each tweet in this dataset is public and contains the following post-related information: 

\begin{itemize}
\small
\item tweet id 
\vspace{-0.035in}
\item posting date
\vspace{-0.035in}
\item number of favorites received
\vspace{-0.035in}
\item number of times it is retweeted
\vspace{-0.035in}
\item the url links shared as a part of it
\vspace{-0.035in}
\item text of the tweet including the hashtags
\vspace{-0.035in}
\item geolocation if tagged
\end{itemize} 

A tweet may contain more than a single hashtag. From this set of tweets, we remove the redundant tweets that are attached to more than one of these four hashtags considered. This resulted in a dataset of 0.2 million tweets that are unique and are posted by a unique set of 33K users. Due to the download limit of the Twitter API, all the tweets in our dataset are from February 2017. None of the tweets we crawled were tagged with a geolocation.

\begin{table}[h]
\centering
\small
\begin{tabular}{ l | c | c }  %\hline
%\hline
\textbf{Metric} & \textbf{AIT} & \textbf{EAIT} \textbf{} \\ \hline %\textbf{Reddit} \\ \hline
\textit{Mean} & 11.033 (86.9) & 11.10 (84) \\ \hline %66.6 (352.8) \\ \hline
\textit{Median} & 11.0 (91) & 11.0 (89) \\ \hline %& 31 (177) \\ \hline
\textit{Min.} & 1 (0) & 1 (0) \\ \hline %& 1 (1) \\ \hline
\textit{Max.} & 30 (140) & 52 (140) \\ \hline %& 3761 (22579) \\ \hline
\end{tabular}
\caption{Statistics about the number of words in a post (and number of characters in a post) crawled from Twitter for AIT vs EAIT}
\label{tab:dataStats}
\end{table}

\subsection{Expert AI-Tweeters (EAIT)}
We manually compile a list of AI experts whom we consider as a seed set to crawl the expert Twitter users. From this seed set, we crawl their friends (users they are following) who are also experts in AI. Through the snowballing approach on the friends list, we compile the EAIT list that contains 9851 expert users. Using the user biography, we label a given user as a EAIT based on these two conditions: 
\begin{enumerate}
\item No vocabulary related to politics, business, news media mentioned -- for example, \emph{reporter}, \emph{organization}, \emph{marketing}, \emph{blockchain}, \emph{breaking}, etc;
\item Vocabulary related to AI is used -- for example, \emph{machinelearning}, \emph{ai}, \emph{vision}, \emph{researcher}, \emph{\#ai}, etc.
\end{enumerate}
%if there are no vocabulary related to politics, business, news media but vocabulary related to AI is used in the user's Twitter biography. 

\noindent
The vocabulary used in this labeling is composed by leveraging the AI vocabulary compiled here: https://goo.gl/ApCbnu. By utilizing the seed set of 100 EAITs, we finally obtain a list of 9851 users. We then use a keyword-based approach to classify this set of users. This classification reveals that 35\% of EAIT are \emph{industry professionals}, 10\% are \emph{academicians}, 6\% are \emph{students} and rest are unclassified. For example, if a user mentions the keyword `student' as part of their biography, that user is categorized as a \emph{student}. To categorize a user as an \emph{academician}, we search for keywords such as `professor', `faculty', `lecturer', `teacher', etc. Similarly, we label an expert user as an industry professional if the Twitter biography contains some of the keywords such as `engineer', `scientist', `director', `developer', `founder', etc. 0.06\% of the tweets we crawled are tagged with geolocation.

\begin{figure*}[!ht]
\centering
\includegraphics[height=3.6in]{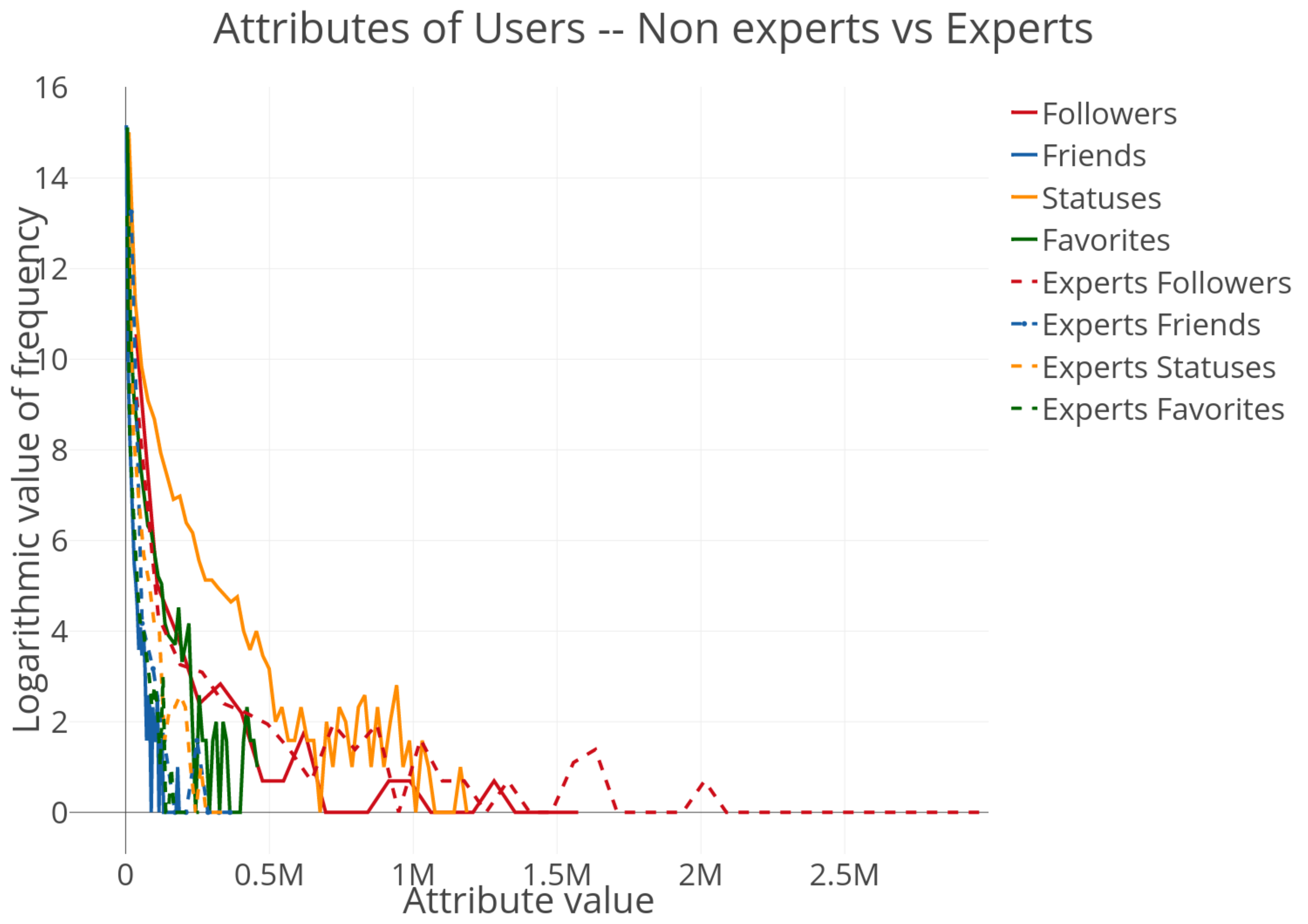}
\caption{User attributes -- Followers; Friends; Statuses; Favorites. The plain thick lines correspond to the AIT and the dotted lines correspond to EAIT. X-axis represents the attribute's value where as, Y-axis represents the logarithmic value of the frequency (Best seen in color)}
\label{fig:userattr}
\end{figure*}

\section{Characterization of Users}
\subsection{Influence attributes}
Before we delve deep to investigate the research questions, we present few details about the demographics of users from AIT and EAIT. To understand the differences between the two types of users, we first focus on the influence attributes. The influence attributes we consider are -- \#statuses shared, \#followers, \#friends, \#favorites. These attributes provide a useful perspective about the activity of users and how their tweets are influencing other Twitter users. To study these attributes, we plot the logarithmic frequencies of these four attributes for AIT and EAIT in Figure~\ref{fig:userattr}. 

Figure~\ref{fig:userattr} shows that from the perspective of sharing and favoriting tweets, both sets of users are active on Twitter but the general users post relatively higher percentage of tweets than EAIT. AIT share more number of statuses than favoriting other tweets. However, in certain cases as shown in Figure~\ref{fig:userattr}, EAIT favorites more or less the same number of tweets as sharing the tweets. When we consider the other influence attributes -- followers and friends, on an average both sets of users have large number of followers than friends (users you are following). EAIT has relatively larger number of followers compared to AIT. The posts shared by AIT are relatively larger than the number of followers they have. In contrast, for EAIT, the number of followers are many times larger than the statuses they share. Existing literature~\cite{Kwak2011for,kwak2012more} shows that the number of retweets and favorites are correlated with followers. This plot shows that the expertise of a user in AI is directly proportional to the number of followers. 

\begin{table}[h]
\sffamily
\scriptsize
\centering
\setlength{\tabcolsep}{3pt}
\begin{tabular}{ | c | p{5.4cm} |} \hline
\textbf{User Category} & \textbf{Geographical Locations} \\ \hline
AIT & USA (3.4\%), India (2.8\%), CA (2.6\%), France (2.4\%), England (1.9\%), NY (1.8\%), UK (1.6\%), London (1.4\%), Germany (0.9\%), Paris (0.9\%) \\ \hline
EAIT & CA (9.7\%), NY (4.5\%), USA (3.2\%), England (2.8\%), France (2.7\%), MA (2.2\%), UK (2.1\%), London (1.9\%), SF (1.8\%), Germany (1.7\%) \\ \hline
\end{tabular}
\caption{Top-10 locations extracted from the user biographies who specified their geographical location}
\label{tab:locations}
\end{table}

Table~\ref{tab:dataStats} compares the statistics about the length of posts made by AIT and EAIT. On an average, tweets made by EAIT are longer compared to the tweets posted by AIT. We then focus on the particulars of the users' professional background and geographical location that are obtained from their profile biographies. 25.1\% of users who are categorized as AIT did not provide their geographical location as part of their profile. Where as, only 15.5\% of users categorized as EAIT did not state their geographical location. Table~\ref{tab:locations} shows that for the AIT category, in the top-10 specification of the locations, 6.77\% are from Europe, 7.74\% of users are from United States (14\% more number of users than from Europe) and  2.8\% are from India. For the EAIT category, 11.17\% from Europe, 21.4\% of users are from United States (almost 91.5\% more number of users compared to Europe) and 0\% from India. Large percentage of experts are from Europe and United States where as large percentage of non-experts talking about AI happen to be from India. 

\begin{table}[h]
\sffamily
\scriptsize
\centering
\setlength{\tabcolsep}{3pt}
\begin{tabular}{ | c | p{5.4cm} |} \hline
\textbf{User Category} & \textbf{Occupation} \\ \hline
AIT & manager, entrepreneur, consultant, founder, developer, engineer, writer, author, blogger, strategist \\ \hline
EAIT & scientist, student, researcher, engineer, professor, cofounder, ceo, founder, director, entrepreneur \\ \hline
\end{tabular}
\caption{Top-10 occupations extracted from biographies}
\label{tab:professions}
\end{table}

We conduct a unigram-based analysis of the profiles of the users to obtain their professional background. Table~\ref{tab:professions} shows that based on the frequencies of professions stated by users on Twitter, majority of the Twitter users contributing to AI-related tweets are pursuing careers in technology.

\subsection{Overall Topics of Interest}
\label{sec:userana}
Since we are characterizing the users, to examine their interests in general, we crawl the most recent 100 posts shared by these users. These posts may talk about AI and non-AI topics. We believe that extracting latent topics over the users' timeline irrespective of the type of post can help measure the level of users' interest in technology that can indirectly acts as a metric to understand their perceptions about AI. We extract the topics from the posts made by users using the Twitter LDA package~\cite{Zhao2011Traditional}. 

\subsubsection{AIT:}
\begin{figure}[ht!]
\centering
\includegraphics[width=\columnwidth]{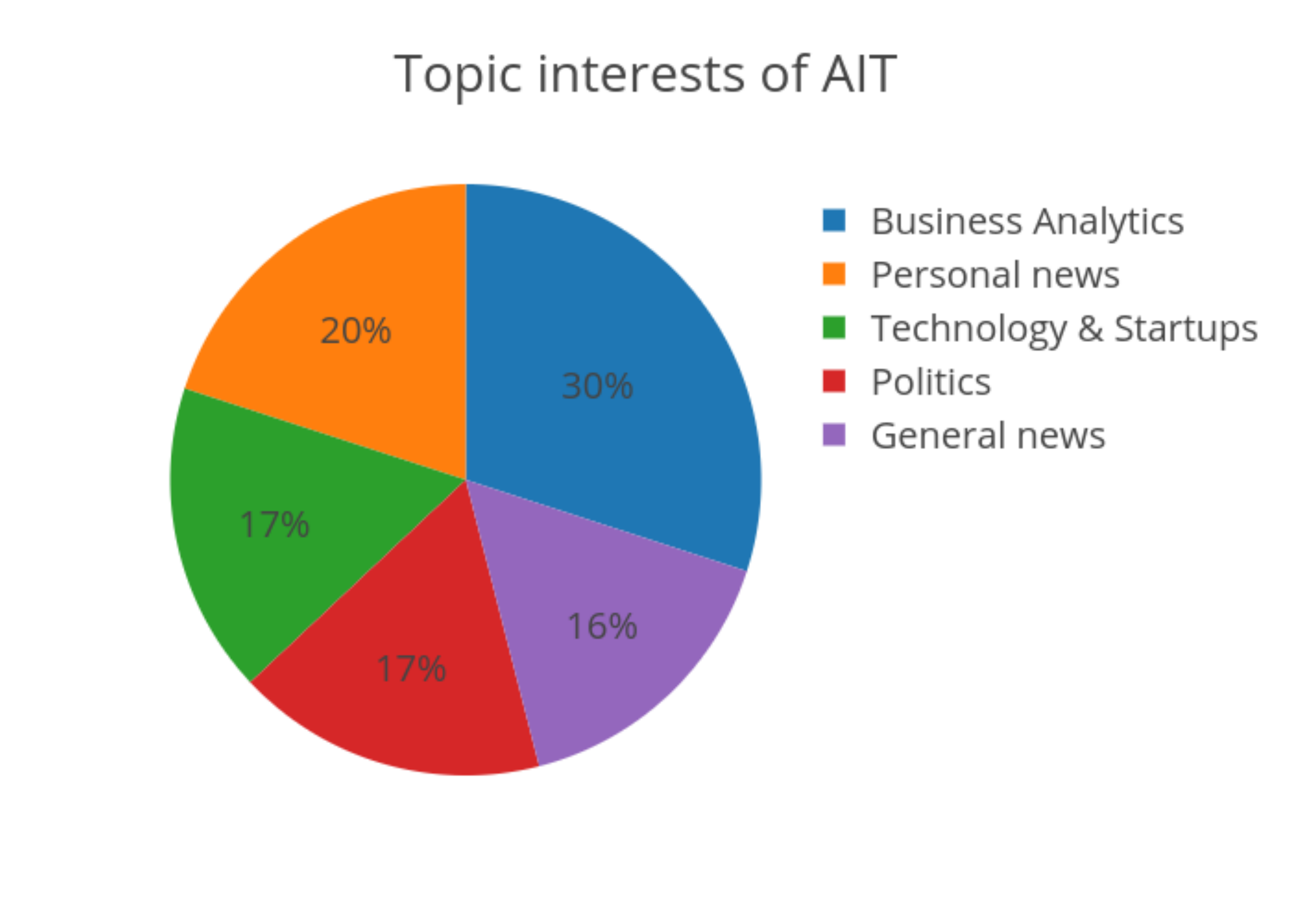}
\caption{Topic Distributions extracted from tweets posted by AIT}
\label{fig:twittopne}
\end{figure}

As mentioned earlier, there are a total of 33K unique set of users who contributed towards our dataset. We crawl their recent tweets to extract the topics. We empirically decided to extract 5 topics and their percentage distribution for each individual's tweets. We then aggregate all the distributions of users across these topics and the percentage distributions are shown in Figure~\ref{fig:twittopne}. The topic distributions show that  large percentage of users are interested in business analytics and then share 20\% of the average number of tweets made by AIT are about their personal news.

%The topics we extracted are inline with the current literature~\cite{LydiaICWSM2016} that there is a considerable percentage of tweets about personal statuses and news. 

%As the existing literature~\cite{LydiaICWSM2016} shows it could also be the case that users are posting their reviews about certain technological gadgets. 
%We could use these findings to state that users who are interested in AI and technology are contributing towards our dataset that we analyzed to measure the public perception. 

\subsubsection{EAIT:}
\begin{figure}[ht!]
\centering
\includegraphics[width=\columnwidth]{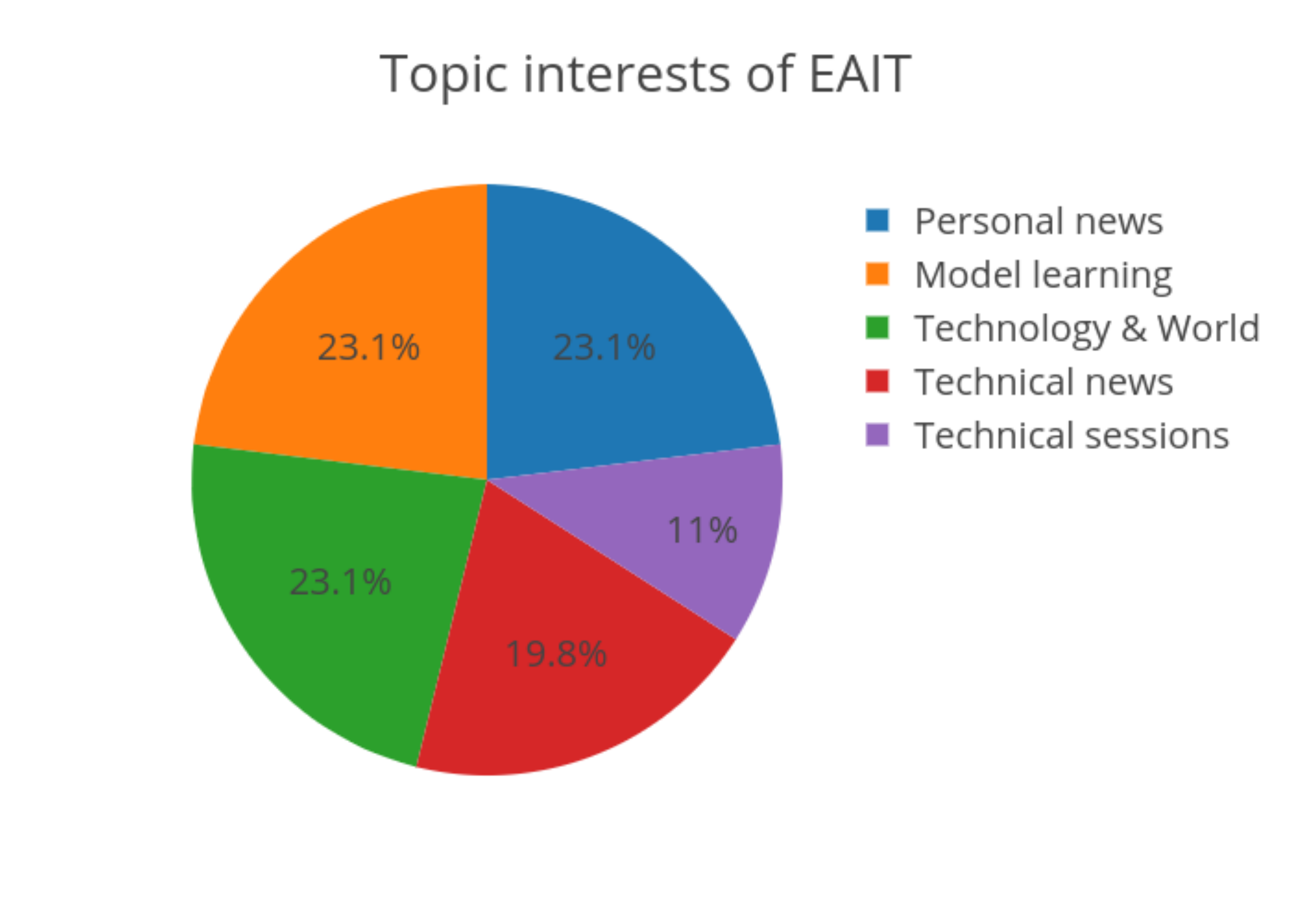}
\caption{Topic Distributions Extracted from tweets posted by EAIT}
\label{fig:twittope}
\end{figure}

%Similar the investigation on Twitter users about the types of topics these users focus on, 
We conduct a similar investigation on the tweets posted by experts. These topics reveal that experts post equal percentages of posts about their personal news, technical implementations of AI systems. These topics are different from the topics focused by AIT.
% This may provide alternate explanation for why tweets posted by the experts receive more favorites than retweets. 
%However, we agree that all AI experts may not necessarily talk only about AI and its related topics. 
The pie chart shown in Figure~\ref{fig:twittope} reveals that more than 77\% of the tweets posted by experts on Twitter are about technology. However, {\em EAIT share significant percentage of personal opinions and statuses where as AIT post the least percentage of tweets about technology}. 

\section{Twitter Engagement}
%In the recent times, there is a significant interest in AI from different technical communities including transportation~\footnote{https://goo.gl/nyY5zi}, healthcare~\footnote{https://goo.gl/s8thiE}, relationships~\footnote{https://goo.gl/e7ubPp}, etc. This interest paved a way to important debates, discussions and birthing of new institutes to study the benefits and risks of AI. However, there is no existing study about how the common users use social media platforms to express their thoughts about AI.

We seek to study the attributes that disclose the holistic picture of the overall engagement rate of AI-related tweets. We believe that the engagement rate has a potential to provide us with the necessary information on the patterns of public interests and perceptions in AI. We measure the engagement by considering the \emph{favorites} received by a tweet, \emph{replies} to a tweet and \emph{mentions} of a Twitter post. The favorites here are the number of likes received by a tweet posted by a user who belongs to either AIT or EAIT. Note that these favorites are different from the favorites we considered as the influence attribute in Section 3.1. We first compute Twitter engagement statistics that are shown in Table~\ref{tab:usereng}.

\begin{table}[!h]
\small
\centering
\setlength{\tabcolsep}{4pt}
\begin{tabular}{ |l|*{2}{c|}*{2}{c|}} \hline
 & \multicolumn{2}{c|}{\emph{Min (Max)}} & \multicolumn{2}{c|}{\emph{Median (Mean)} } \\ \hline 
 & \textbf{AIT} & \textbf{EAIT} & \textbf{AIT} & \textbf{EAIT} \\ \hline
\textbf{Retweets} & 0 (1041) & 0 (1701) & 0.0 (1.5) & 0.0 (3.28)\\ \hline
\textbf{Favorites} & 0 (1268) & 0 (1914) & 0.0 (1.46) & 1.0 (4.98)\\ \hline
\textbf{Mentions} & 0 (9) & 0 (10) &  0.0 (0.63) & 0.0 (0.54)\\ \hline
\end{tabular}
\caption{Min (Max) and Median (Mean) values of Retweets, Favorites, Mentions}
\label{tab:usereng}
\end{table}

%For the Twitter users category, t
Tweets made by EAIT are more likely to be retweeted than favorited by the users on this platform. 71.93\% of EAITs tweets are retweeted atleast once and 31.14\% of tweets are favorited atleast once. The percentage of EAIT's posts retweeted is significantly higher than the general dataset which is 11.99\% as shown by the existing literature~\cite{Suh2010socom}. Where as, average percentage of tweets posted by AIT retweeted (69.38\%) atleast once is almost same as the average percentage of favorites (67.8\%) received per tweet. 

Tweets from AIT has 11.45\% of tweets that contain atleast one user handle where as, EAIT has 67.57\% of such tweets. This shows that experts are more likely to interact or engage in discussions with each other about AI on Twitter than AIT. Literature~\cite{Java2007,Kwak2010,yang2010predicting} considers retweeting as one of the features to measure information diffusion. Based on these results, {\em tweets posted by EAIT diffuse faster (higher retweet rate) than the tweets posted by AIT}.

\begin{figure*}[!ht]
        \centering
    \begin{subfigure}[t]{0.3\textwidth}
        \centering
        \includegraphics[height=1.2in]{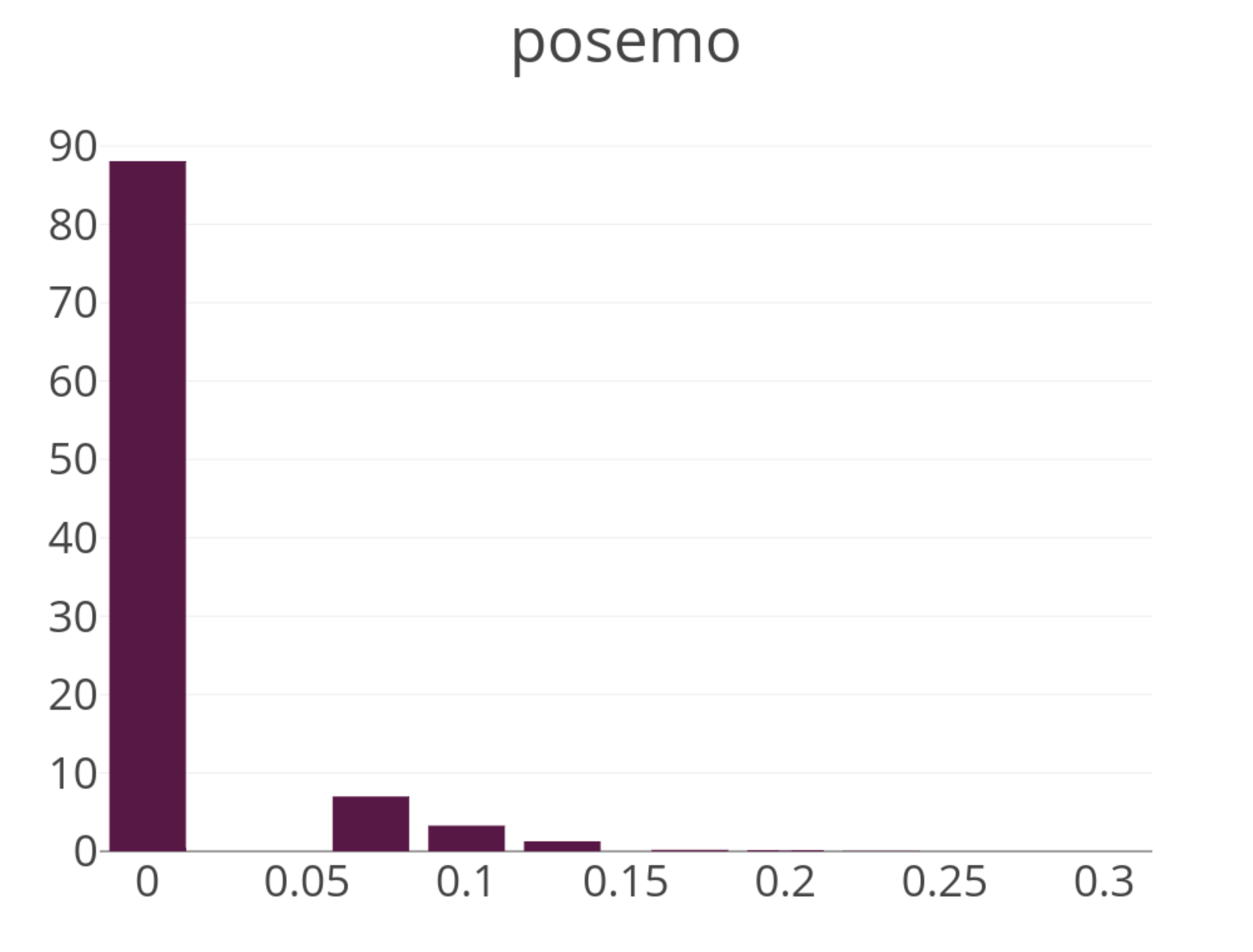}
	\caption{Positive Emotions}
    \end{subfigure}%
    ~ 
    \begin{subfigure}[t]{0.3\textwidth}
        \centering
        \includegraphics[height=1.2in]{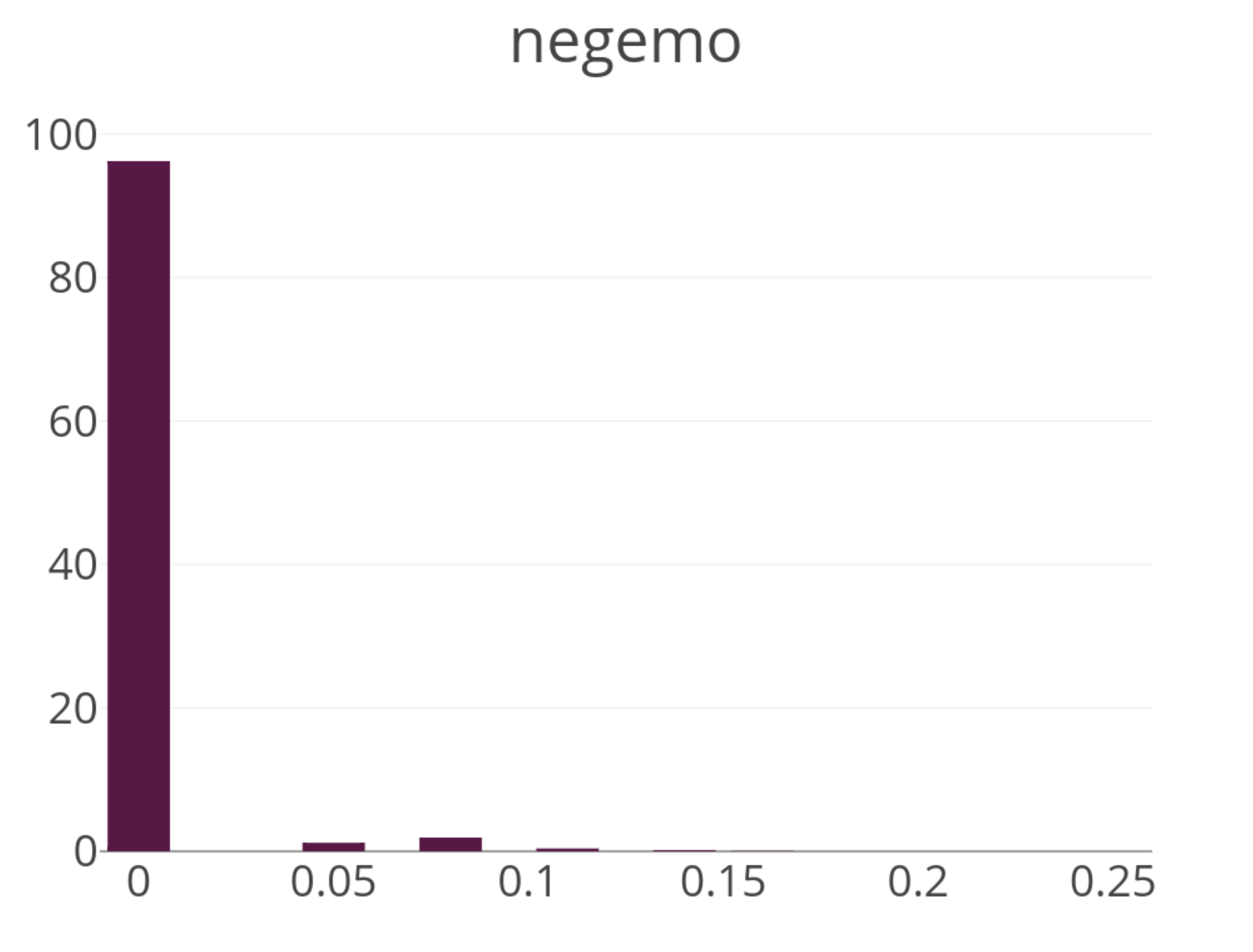}
	\caption{Negative Emotions}
    \end{subfigure}
    ~ 
    \begin{subfigure}[t]{0.3\textwidth}
        \centering
        \includegraphics[height=1.2in]{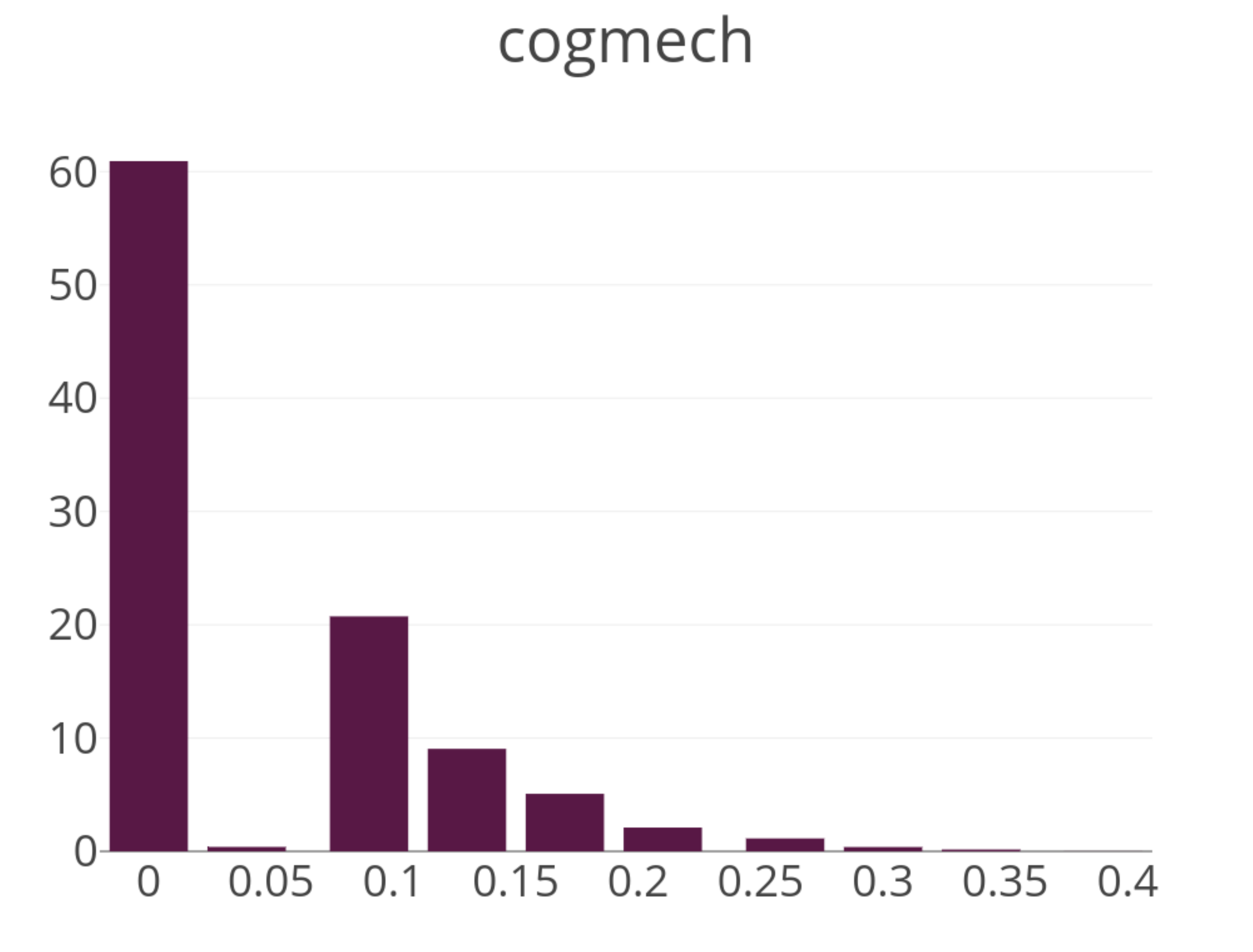}
	\caption{Cognitive depth}
    \end{subfigure}
    \caption{Emotions from the tweets posted by AIT}
    \label{fig:liwcnexp}

\end{figure*}

\section{Optimistic or Pessimistic}
%Optimism is defined as being hopeful and confident about the future whose synonym is `positive' and pessimism is defined as tending to see or believing that the worst will happen whose synonym is `negative'. 
In this work we measure the emotion attributes -- optimism and pessimism in terms of positive and negative emotions. Alongside we assess other emotion-related attributes like cognitive mechanisms, insights and social aspects. Tausczik et. al~\cite{TausczikY2010} in their work introducing LIWC mention that the way people express emotion and the degree to which they express it can tell us how people are experiencing the world. Existing literature~\cite{Danescu2011words,Tsur2012wsdm,Tumasjan2010Twitter} states that LIWC is powerful in accurately identifying emotion in the usage of language. Considering this fact, we employ the psycho-linguistic tool LIWC to measure the emotionality expressed in the tweets.

%\cite{TausczikY2010} describes that usage of insight words like think, know, consider, etc., and causal words like because, effect, hence, etc., suggests the active reassessment of something or someone and to create causal explanations is to organize the thoughts of an individual respectively. The same article also describes that usage of conjunctions and cognitive mechanisms help join multiple thoughts together to create a coherent narrative and can be thought of as richness in reasoning. Work and social attributes measured by using LIWC suggests the personality of individuals who are high on extroversion. 

\subsection{AIT}
Figure~\ref{fig:liwcnexp} reveals that users categorized as AITs are more positive (65\% greater than negative) and optimistic towards AI and its related topics. The horizontal axis in this figure represents the value of a given emotion attribute and the frequency of that attribute is plotted on the corresponding vertical axis. The distribution in each plot are normalized and the sum of all the values in different buckets of emotion metric will sum up to 100\%. These results concur with the recent literature~\cite{EthanAAAI2016,LeslieHarvard2016} on New York Times articles and interviews with individuals. This is a useful finding because prior work shows that Twitter is known for more emotionally negative~\cite{LydiaICWSM2016} posts. In other words, {\em despite the general negative emotional content on Twitter, this subset of tweets focusing on artificial intelligence are more positive than being negative}.

\begin{figure*}[!ht]
\centering
    \begin{subfigure}[t]{0.3\textwidth}
        \centering
        \includegraphics[height=1.2in]{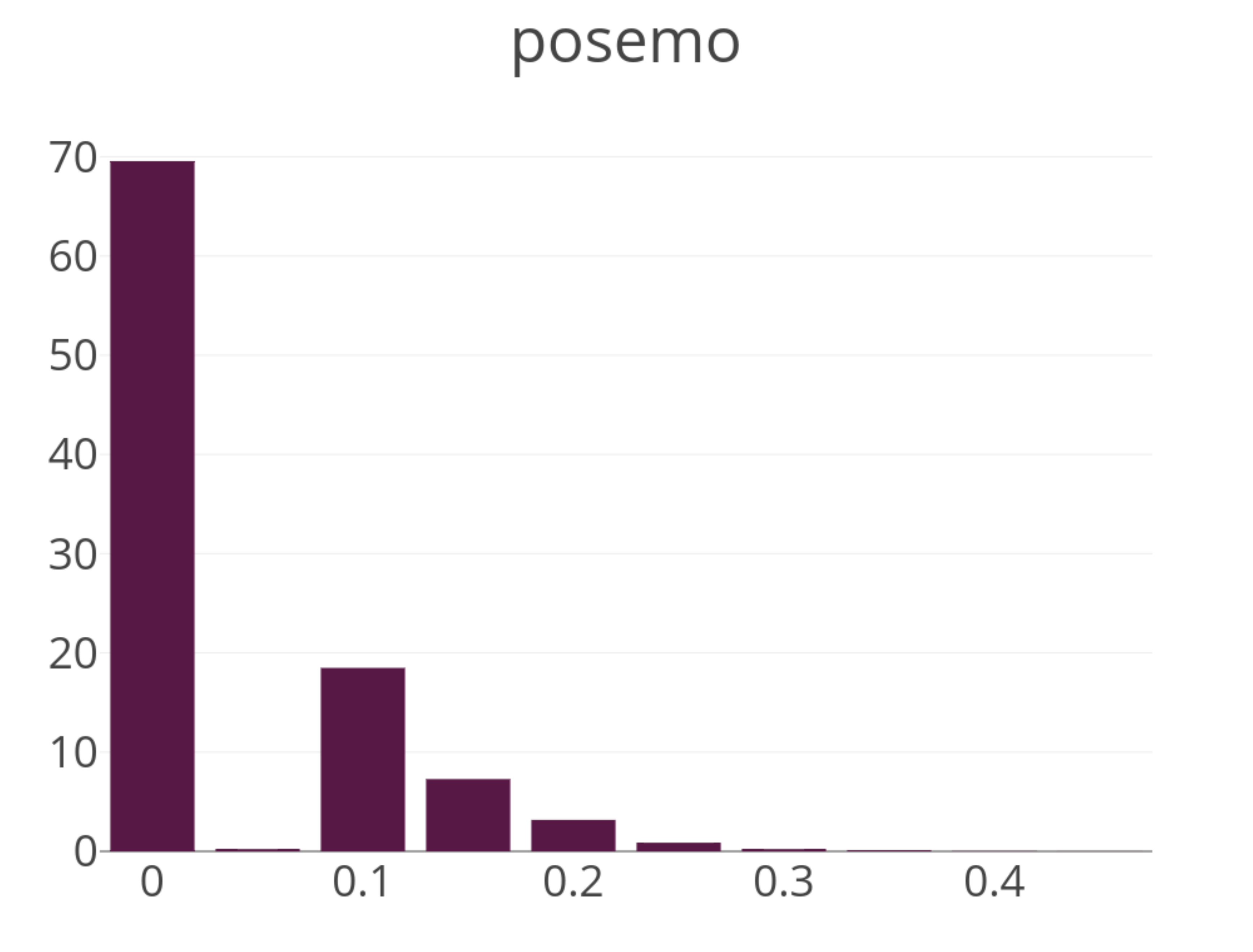}
	\caption{Positive Emotions}
    \end{subfigure}%
    ~ 
    \begin{subfigure}[t]{0.3\textwidth}
        \centering
        \includegraphics[height=1.2in]{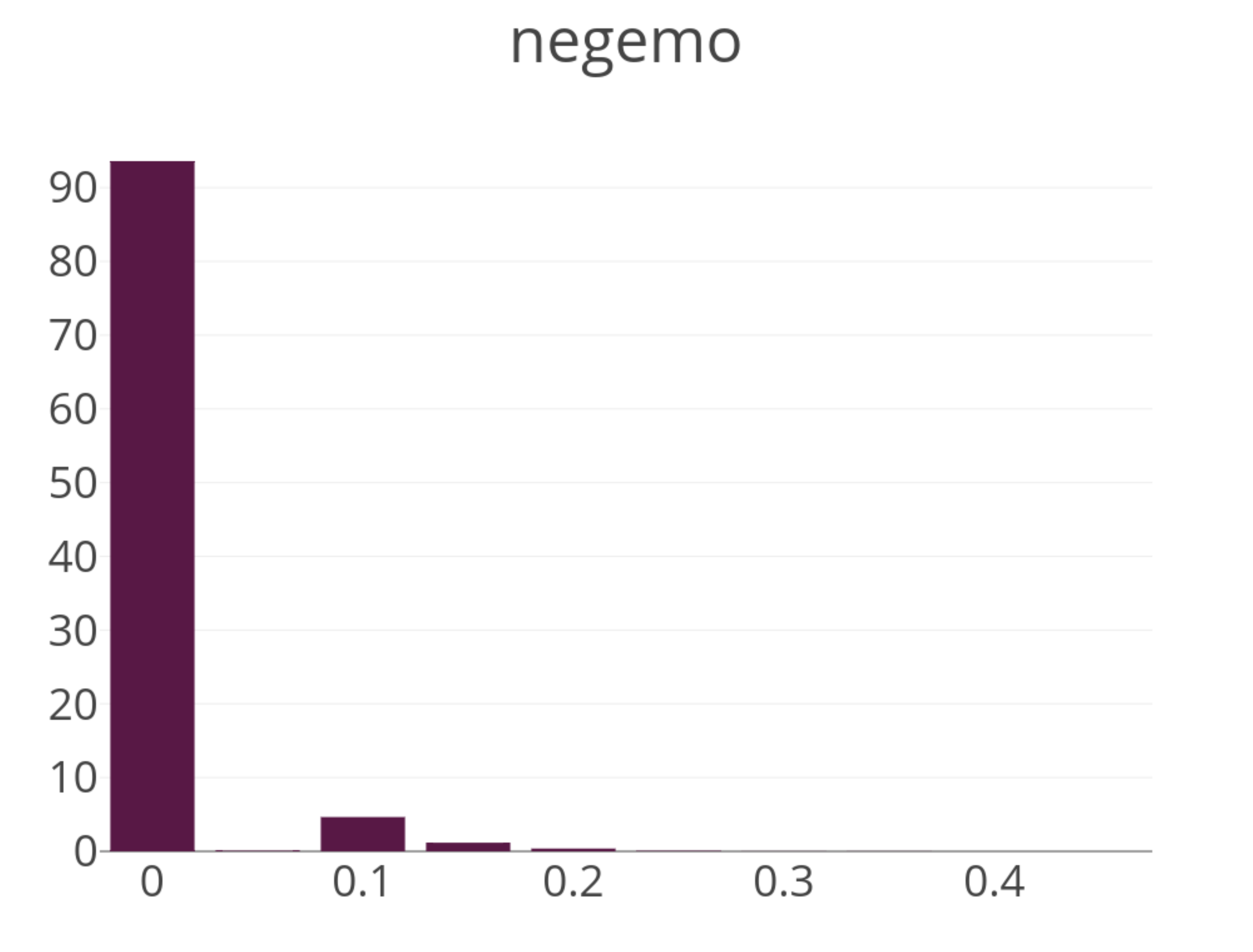}
	\caption{Negative Emotions}
    \end{subfigure}
    ~ 
    \begin{subfigure}[t]{0.3\textwidth}
        \centering
        \includegraphics[height=1.2in]{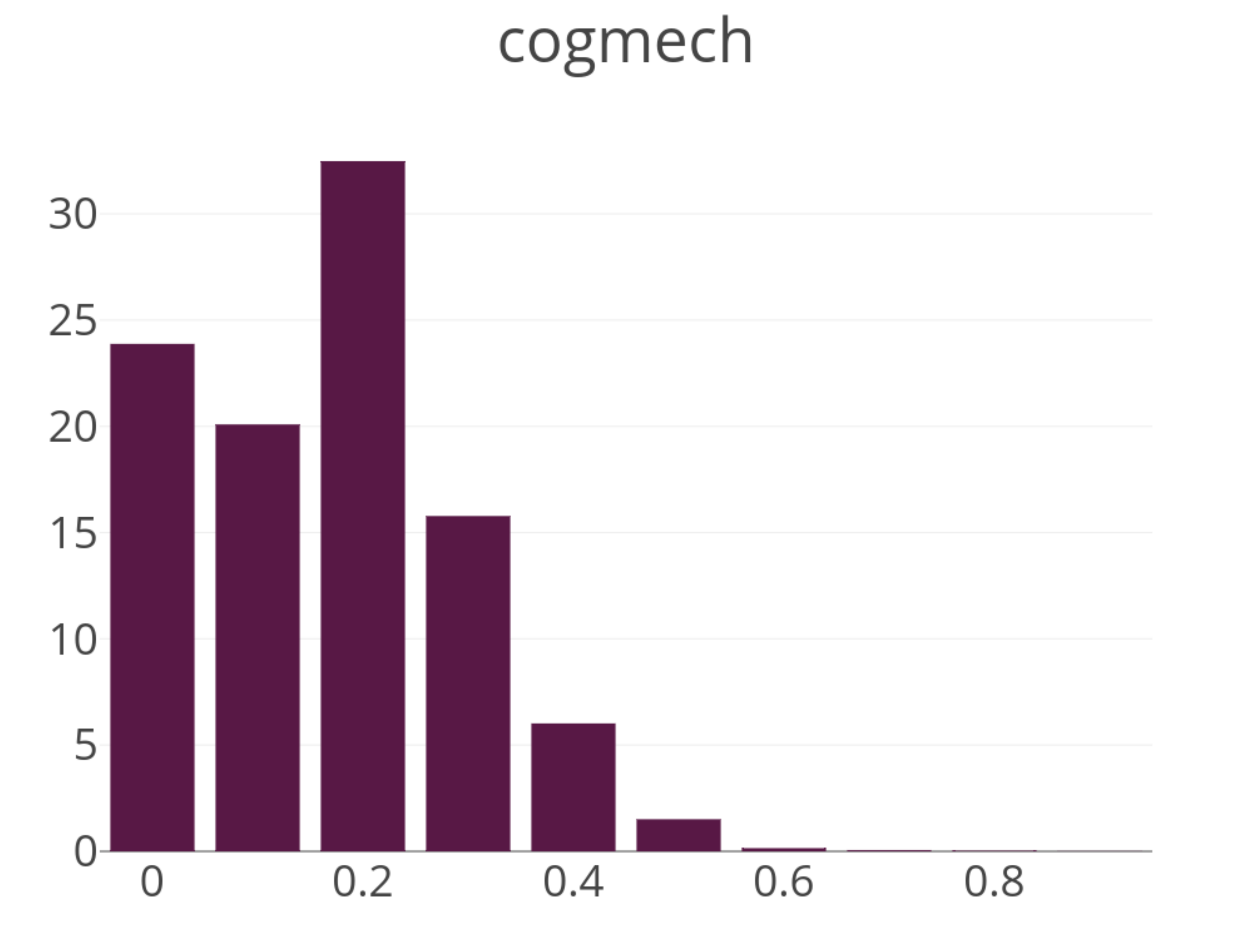}
	\caption{Cognitive depth}
    \end{subfigure}
    \caption{Emotions from the tweets posted by EAIT}
    \label{fig:liwcexp}
\end{figure*}

\subsection{EAIT}
We conduct the similar emotion analysis on tweets posted by experts and it reveals similar findings as earlier (shown in Figure~\ref{fig:liwcexp}) but with relatively higher negativity and higher cognitive mechanisms compared to AIT. Cognitive mechanisms or complexity can be described as richer way of reasoning~\cite{TausczikY2010}. When certain set of posts have high values of cognitive mechanisms, this means that this set of posts contain large percentage of technical content. The horizontal axis in Figure~\ref{fig:liwcexp} represents the gravity of a given emotion and the vertical axis represents the number of posts with the corresponding  value of the given emotion on horizontal axis. 
Compared to AIT, tweets made by EAIT have more negativity overall. However, positive emotion is four times as dominating as the negative emotion. When we compare the positive and negative emotions of the two categories of users, {\em the results reveal that expert users (\emph{pos-index}:3.25; \emph{neg-index}: 0.60) are almost twice the percentage of being negative than the AIT (\emph{pos-index}:0.82; \emph{neg-index}: 0.248)}.

Alongside, we conduct a granular evaluation by comparing the metrics of emotion between the three sub-categories of expert users -- students, academicians and industry professionals. The aggregated values shown in Table~\ref{tab:finelevelemo} suggest that academicians are relatively less positive and more social than the users from the other two categories when tweeting about AI. These results corroborate the results from Twitter engagement where experts are using relatively larger percentage of mentions in their tweets engaging in discussions with others. These results also show that students and industry professionals tweet relatively more insights about AI in their tweets than academicians. According to Tausczik et. al~\cite{TausczikY2010}, insight words suggest the active reassessment of a theory.  

\begin{table}[!h]
\small
\centering
\setlength{\tabcolsep}{4pt}
\begin{tabular}{ |l|c|c|c|c|c|} \hline
\emph{User Type} & \emph{PA} & \emph{NA} & \emph{COG} & \emph{INSG} & \emph{Soc} \\ \hline
\hline
\emph{Students} & 3.14 & 0.70 & 23.06 & 13.91 & 3.77 \\ \hline
\emph{Academicians} & 2.72 & 0.70 & 21.60 & 12.84 & 3.85 \\ \hline
\emph{Industry Prof.} & 3.19 & 0.60 & 22.92 & 13.84 & 3.13 \\ \hline
\end{tabular}
\caption{Aggregated values of different metrics of emotion-- Positive Affect (PA); Negative Affect(NA); Cognitive depth (COG); Insights (INSG); Social aspects (Soc) for three categories of experts -- students; academicians; industry professionals}
\label{tab:finelevelemo}
\end{table}

\section{Topics heavily discussed by users on Twitter about AI}
In Section~\ref{sec:userana}, we have presented the analysis on the interests of users by crawling their timelines and extracting topics from their timelines. In order to better understand the public perceptions about AI, we extract topics from the AI-related tweets. In this section we focus only on the AI-related tweets posted by users from AIT and EAIT. To perform this, we first consider all the tweets posted by AIT and EAIT separately. We utilize a \emph{keyword}-based approach that looks for specific AI-related vocabulary in any given tweet. As there is a possibility that some tweets might be retweeted by the same set of users in a given category, we pre-process the two sets of AI-related tweets from AIT and EAIT. Once we clean the data, we then combine these two datasets to identify the latent topics. We empirically chose 6 topics to avoid redundancy and then identify the percentage of tweets that are contributing to each topic from these two categories -- AIT and EAIT.  

In Table~\ref{tab:aivocab}, we present the topics extracted from the AI-related tweets. These topics display that the largest percentage of tweets shared by AIT (37\%) focus on the effects of automation on future. Where as, the largest percentage of tweets made by EAIT (25\%) concentrates on the technical implementations of AI systems. This topic is followed by tweets focusing on conferences \& talks related to AI. This could be explained due to the interests of the crowd that is considered as experts in this analysis. The emphasis on the applications of AI from industry are relatively equal among both AIT and EAIT. As expected, the results show that AIT focus on general news about AI and the myths associated with AI more than the expert users. Due to the partial alignment of these topics with the findings shown by Fast et. al~\cite{EthanAAAI2016}, individuals have continued interests in the similar topics over years. 

\begin{figure}[!ht]
\centering
\includegraphics[width=\columnwidth]{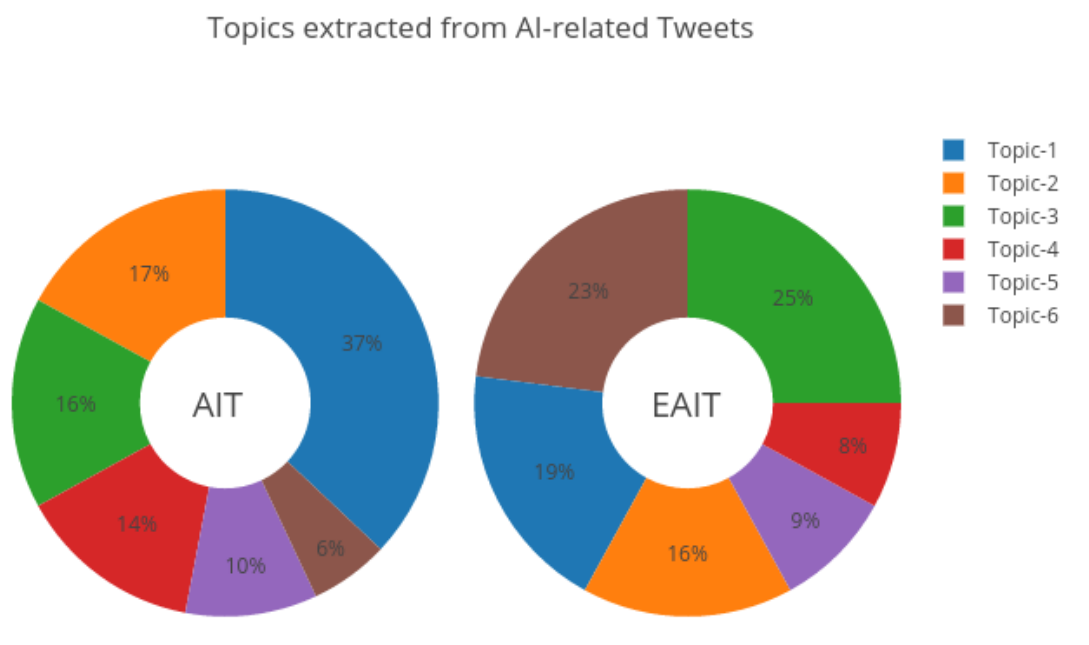}
\caption{Topics (shown in Table~\ref{tab:aivocab}) extracted from the AI-related tweets}
\label{fig:AITopics}
\end{figure}

\begin{table*}[!ht]
\centering
%\resizebox{\textwidth}{!}{
%\footnotesize
%\small
\begin{tabular}{l | l |  p{7.4cm} }  %\hline
\textbf{ID} & \textbf{Topic} & \textbf{Top Tags}  \\ \hline
1 & Effects of automation on future & future, business, human, jobs, revolution, automation, experience, impact, change, improve \\ \hline
2 & AI applications from Industry & humans, google, elon, robots, facebook, brain, deepmind, cars, selfdriving, bots \\ \hline
3 & Technical aspects of building models & learning, data, deep, analytics, algorithms, python, models, cloud, model, training \\ \hline
4 & Daily news & latest, daily, news, tech, assistant, cars, mobile, voice, robot, speech, alexa \\ \hline
5 & Myths \& rise of AI & data, myths, automation, language, internet, age, machines, rise, language, read \\ \hline
6 & Conference News & learning, talk, join, workshop, conference, event, meetup, summit, talking, panel, session \\ \hline
\end{tabular}
\caption{Topics and their corresponding vocabulary extracted from the AI-related tweets.}
\label{tab:aivocab}
\end{table*}

% EAIT makes posts about AI which we have seen in the subsection~\ref{sec:userana}. AIT is mainly interested in discussing about the stories and statistics of AI along with deep learning. Where as, EAIT is mainly engaged in debating about the rise of AI and sharing the news about research breakthroughs. These findings especially the shallow interests of AIT in AI are inline with the current literature~\cite{}.

\section{Co-occurring concepts}
The questions we investigated until now provides valuable insights into whether and how individuals perceive the issues about AI advancements. However, we note that conceptual relationships could significantly quantify and measure the perceptions of individuals. Towards addressing this challenge, we employ the popular word2vec analysis to detect relationships between words that are frequently co-occurring. Word2Vec~\cite{TomasWord13} is a popular two-layer neural network that is used to process text. It considers a text corpus as an input and generates feature vectors for words present in that corpus. Word2vec represents words in a higher-dimensional feature space and makes accurate predictions about the meaning of a word based on its past occurrences. These vectors can then be utilized to detect relationships between words which are highly accurate given enough data to learn these vectors. 
\begin{table*}[ht]
\begin{tabular}{c|c|c|c|c|c|c|c}
\toprule
\multicolumn{2}{c}{\textbf{\em Agents}} & \multicolumn{2}{|c|}{\textbf{\em Robots}} & \multicolumn{2}{c}{\textbf{\em Ethics}}& \multicolumn{2}{|c}{\textbf{\em Privacy}}\\ \hline
%\cmidrule{1-2} \\%& \cmidrule{3-4} & \cmidrule{5-6} \\ 
\emph{AIT}& \emph{EAIT}&  \emph{AIT}& \emph{EAIT}& \emph{AIT}&\emph{EAIT} & \emph{AIT} & \emph{EAIT} \\ \hline
%agents & agents & robots & robots & ethics & ethics & privacy& privacy\\ 
aiding & inattention & human & smart & empathy &  philosophy & healthcare& guarantee\\ 
actions & intelligence & creative & killer & philosophy & utilitarian  & protecting & challenge\\
intelligence & nonhuman & machines & simulators & moral & morality & robust & healthcare \\
cooperate & aggression & devices & toy & kant & humanism & discrimination & papers \\
egotistical & activities & software & creators & nietzsche & politics & complicates & secrecy\\
humanity & spy & humanoid & machines & considerations & personhood & enforcement & accountability\\
trained & cooperate & designing & tropes & cyber & perspectives & regulations & infringement\\
researchers & brokers & hacking & devices & thoughts & blog & bias & implementing \\
emotional & supervised & worlds & fantastic & journalism & principles & watchdog & challenges \\
kiosks & deflect & rogue & universe & privacy & virtues & cyber & authentication \\
\bottomrule
\end{tabular}
\caption{Top-10 co-occurring words with a given popular keyword; Co-occurrence patterns are extracted using Word2vec model trained with AI-related tweets.}
\label{tab:cooccur}
\end{table*} 

To detect the relationships, we train the Word2Vec model separately on the 71915 and 72,153 AI-related tweets posted by EAIT and AIT respectively. As a processing step, we first remove stop words from the tweets and consider each tweet independently. We utilized the pre-existing lists from academia\footnote{https://www.cs.utexas.edu/users/novak/aivocab.html} and industry\footnote{http://www.techrepublic.com/article/mini-glossary-ai-terms-you-should-know/} to manually compile the AI vocabulary of 61 words. Table~\ref{tab:cooccur} provides the top-10 words co-occurring with the four keywords related to AI -- \emph{agents}, \emph{robots}, \emph{ethics} and \emph{privacy}. These words in the table are sorted in the decreasing order of their co-occurrence probability.

The co-occurring patterns shown in Table~\ref{tab:cooccur} suggest that AIT and EAIT use terms strikingly different. 
\begin{itemize}
\item \emph{Agents} -- It shows that EAIT are focusing on the behavioral characteristics of intelligent agents by using the terms -- \emph{inattention}, \emph{intelligence}, \emph{aggression}, etc. AIT also use similar terms related to behavior but they also focus on the applications of these intelligent agents.
\item \emph{Robots} -- AIT are focusing on the physical aspects of robots and their design issues. However, EAIT associates words that describe the types of robots and their usability. 
\item \emph{Ethics}-- AIT are in general concerned about ethics related to AI especially expressing through words like \emph{empathy}, \emph{moral} and \emph{privacy}. However, from the words used by EAIT it shows that they relate ethics to different aspects of humanities. 
\item \emph{privacy} -- AIT associates privacy with the drawbacks of AI systems -- \emph{bias}, \emph{discrimination}, etc which sound more negative. However, EAIT focuses on the implementation aspects of AI systems which can maintain accountability and respect privacy. 
\end{itemize}

%The entire list of co-occurrence relationships for the 61 words in AI vocabulary will be released after the review results are published for this paper. 

% \begin{figure*}[!ht]
% \centering
%     \begin{subfigure}{0.3\textwidth}
%         \centering
% 	\includegraphics[scale=0.23]{nonexp_agents.pdf}
% 	\caption{Agents--AIT}
%     \end{subfigure}
%     ~
%     \begin{subfigure}{0.3\textwidth}
% 	\centering
% 	\includegraphics[scale=0.23]{nonexp_robots.pdf}
% 	\caption{Robots--AIT}
%     \end{subfigure}
%     ~
%     \begin{subfigure}{0.3\textwidth}
% 	\centering
% 	\includegraphics[scale=0.23]{nonexp_ethics.pdf}
% 	\caption{Ethics--AIT}
%     \end{subfigure}
%     ~
%     \begin{subfigure}{0.3\textwidth}
% 	\centering
% 	\includegraphics[scale=0.23]{exp_agents.pdf}
% 	\caption{Agents--EAIT}
%     \end{subfigure}
%     ~
%     \begin{subfigure}{0.3\textwidth}
%         \centering
% 	\includegraphics[scale=0.23]{exp_robots.pdf}
% 	\caption{Robots--EAIT}
%     \end{subfigure}
%     ~
%     \begin{subfigure}{0.3\textwidth}
% 	\centering
% 	\includegraphics[scale=0.23]{exp_ethics.pdf}
% 	\caption{Ethics--EAIT}
%     \end{subfigure}
%     \caption{Heatmaps of the co-occurrence patterns associated with the 3 important buzzwords of AI -- \emph{agents}, \emph{robots} and \emph{ethics}.}
%     \label{fig:cooccur}
% \end{figure*}

\section{Conclusions}
Social media platforms are one of the primary channels of communication in the lives of individuals. These platforms are reshaping our ideas and the way we share those ideas. Given the increasing interest in AI from different communities, multiple debates are commencing to evaluate the benefits and drawbacks of AI to humans and society as a whole. This paper presents the findings from our investigation on public perceptions about AI using the AI-related posts shared on Twitter. Alongside, we performed a comparative analysis between how the posts made by AIT and EAIT are engaged. Some of the key findings from our analysis are: 

\begin{enumerate}
\item Based on the user characterization analysis, it was revealed that users who post about AI on Twitter are predominantly from USA and Europe.  
\item Tweets about AI are overall more positive compared to the general tweets posted on Twitter. 
\item AI-related tweets posted by EAIT are more negative than the AIT. 
\item The effects of automation are of predominant concern to the general AI tweeters than the experts. 
\item The large percentage of tweets made by the experts are about the technical implementations of AI systems and news about conferences. 
\item Tweets posted by students and industry professionals relatively provide more insights about AI than academicians. 
\item Academicians are relatively less positive and more social than students and industry professionals when tweeting about AI.
\item Tweets posted by experts have higher diffusion than the tweets posted by general AI tweeters. 
\end{enumerate}

The co-occurring pattern mapping tells us that EAIT acknowledges the challenges in building intelligent systems that maintains accountability in terms of honoring ethics and privacy. The terms used by them also focus on the implementation aspects of such systems. Where as, the terms used by AIT refers to their concerns about the behavioral characteristics and drawbacks of AI systems.
%do not focus much on ethics and design issues of robots compared to AIT.  %Additionally, the most discussed topics on Twitter by the general AI tweeters are about the \emph{future effects of automation} along with an emphasis on \emph{non-technical aspects}. However, experts post significant percentage of tweets about their personal news. Emotion analysis also revealed that discussions are cognitively loaded with a larger gravities among expert users.

We hope that our findings will benefit different organizations and communities who are debating about the benefits and threats of AI to our society. Some of the future directions include a longitudinal study across several years as well as multiple mediums of communication. 

\bibliographystyle{aaai}
\bibliography{references}

\end{document}